\documentclass[runningheads]{llncs}

\usepackage{eccv}

\usepackage{eccvabbrv}

\usepackage{graphicx}
\usepackage{booktabs}

\usepackage[accsupp]{axessibility}  

\usepackage{hyperref}

\usepackage{orcidlink}

\begin{document}

\title{AffectFlow-DINO: Uncertainty-Aware Multi-Task Affect Estimation via Conditional Rectified Flow}

\titlerunning{ABAW 2026 Multi-Task Affect Analysis}

\author{Salah Eddine Bekhouche\inst{1} \and
Abdellah Zakaria Sellam\inst{2} \and
Fadi Dornaika\inst{1} \and
Abdenour Hadid\inst{3}}

\authorrunning{S.E.~Bekhouche et al.}

\institute{University of the Basque Country (UPV/EHU), Spain \and
Department of Innovation Engineering, University of Salento \& Institute of Applied Sciences and Intelligent Systems (CNR), Italy \and
Faculty of Data Science and Computing, Universiti Malaysia Kelantan, Malaysia}

\maketitle

\begin{abstract}
We present \textbf{AffectFlow-DINO}, a multi-task learning system for the 11th ABAW challenge that extends a standard deterministic architecture with a conditional rectified-flow head to model the inherent ambiguity of in-the-wild facial behavior. Instead of predicting a single affect estimate, the model learns a conditional generative distribution, enabling uncertainty-aware one-to-many predictions through Monte Carlo sampling. The system jointly estimates continuous valence-arousal, classifies eight facial expressions, and detects twelve Action Units from static face images. Built on a frozen DINOv3 ViT-S/16 backbone, extensive ablation studies show that rectified-flow decoding consistently improves deterministic prediction, particularly for valence-arousal estimation (CCC-V $+0.058$). We further show that post-hoc threshold calibration effectively recovers performance on severely imbalanced rare classes (e.g., Fear: $3.8\% \rightarrow 33.1\%$) without retraining. Combined with backbone fine-tuning and flow retuning, the final model achieves $\mathbf{P_{MTL}=1.177}$, substantially outperforming the official challenge baseline of $P_{MTL}=0.45$.

\end{abstract}

\section{Introduction}
\label{sec:intro}

The 11th Affective Behavior Analysis in-the-wild (ABAW) competition includes a Multi-Task Learning (MTL) challenge in which each facial frame is associated with three complementary affect representations: continuous valence-arousal (VA) values, an eight-way expression label, and twelve binary Action Unit (AU) labels.
This setting is challenging because the three targets differ in output type and supervision density, and because s-Aff-Wild2 exhibits severe class imbalance (an 8$\times$ expression imbalance and a 29$\times$ AU imbalance) while joint three-task annotation covers only 37\% of training frames.

We present \textbf{AffectFlow-DINO}, a frame-level MTL system that augments a DINOv3 ViT-S/16 backbone with a conditional rectified-flow head~\cite{liu2022flow} to model the full joint affect distribution $p(y \mid x)$.
The core motivation is the inherent perceptual ambiguity of in-the-wild facial behavior: a subtle smile, partial occlusion, or low-intensity expression may correspond to several plausible configurations in the joint VA, expression, and AU space, yet deterministic point estimates collapse this ambiguity into a single vector and discard the uncertainty.
The rectified-flow head learns instead a generative transport map from noise to $p(y \mid x)$, producing a \emph{family} of plausible affect vectors per image whose mean improves over the point estimate and whose spread characterizes prediction uncertainty.
Our main contributions are:
\begin{itemize}
  \item A masked rectified-flow MTL objective that jointly optimizes deterministic task heads and a generative conditional distribution over the 22-dimensional affect vector.
  \item A systematic ablation study spanning 26 design choices on s-Aff-Wild2, covering flow loss weight, inference efficiency, imbalance remedies, backbone fine-tuning, and post-hoc calibration strategies.
  \item Post-hoc per-AU and per-class expression threshold calibration, yielding a combined $P_{MTL}=1.177$ on the validation split, a $+0.054$ improvement from calibration alone, driven by recovering Fear F1 from $3.8\%$ to $33.1\%$ and Sadness from $17.1\%$ to $28.2\%$.
\end{itemize}

\section{Related Work}
\label{sec:related_work}

\paragraph{ABAW and Aff-Wild lineage.}
The Aff-Wild/Aff-Wild2 benchmark series introduced in-the-wild dimensional affect prediction and progressively combined the three core targets, continuous valence-arousal (VA), categorical expressions, and binary Action Units (AUs), into a joint multi-task challenge~\cite{zafeiriou2017aff,kollias2019deep,kollias2019expression,kollias2021affect}. ABAW1--6 built on this foundation, expanding dataset scale, task coverage, and evaluation protocols~\cite{kollias2020analysing,kollias2021analysing,kollias2022abaw,kollias2023abaw,kollias2023abaw2,kollias20246th}; the 7th ABAW competition then established the static-frame, three-task MTL format with the summed metric $P_{MTL}$ used in the current 11th ABAW challenge~\cite{kollias20247th}. ABAW8--10 expanded toward multimodal and fine-grained behavior analysis, and ABAW11 continues this trajectory while retaining the VA+EXPR+AU MTL track~\cite{kollias2025advancements,kollias2025emotions,kollias2026affectcomplexbehavioradvancing}.

\paragraph{Datasets.}
s-Aff-Wild2~\cite{kollias2024behaviour4all} is the benchmark for this challenge; related datasets (AffectNet, RAF-DB, DFEW, EmotioNet, BP4D, DISFA, and others) cover partial subsets of the task triad and are summarized in Appendix~\ref{app:datasets} (Table~\ref{tab:related_datasets})~\cite{mollahosseini2017affectnet,li2017reliable,jiang2020dfew,wang2022ferv39k,benitez2017emotionet,zhang2014bp4d,mavadati2013disfa,kossaifi2017afewva,kossaifi2019sewa,barros2018omg,kollias2023cexpr}.

\paragraph{MTL for affect.}
Early MTL models showed that VA, expressions, and AUs are complementary representations learnable via shared encoders~\cite{kollias2019face,kollias2021affect,kollias2021distribution,kollias2024distribution}. Practically, label incompleteness, task-structure heterogeneity (regression vs.\ classification vs.\ multi-label detection), and negative transfer between tasks require per-task masking, selective fusion, and carefully tuned loss weights~\cite{liu2019mtl}. Recent ABAW solutions combine strong pretrained backbones (MAE, DINOv2) with task-aware fusion, staged training, and temporal modeling~\cite{he2022mae,oquab2023dinov2,liu2024progressive,cabas2024ddamfn,li2024taskadaptive,yu2025solution}.

\paragraph{Generative and uncertainty-aware prediction.}
Diffusion and flow-matching models applied to structured prediction have shown that generative transports over label space can outperform deterministic regressors when targets are inherently multimodal; in affective computing, uncertainty in VA estimation has been addressed via Gaussian heads and evidential regression, but these scalar-variance extensions cannot capture multimodal ambiguity across the joint affect space.
Conflict-aware fusion approaches specifically highlight that ambivalent facial behavior, where modality cues disagree, calls for richer predictive models than point estimates~\cite{bekhouche2026conflict}.
Rectified flow~\cite{liu2022flow} offers an elegant solution: a straight-line transport from noise to data that enables fast few-step inference over the joint 22-dimensional affect target, to our knowledge the first such application in a heterogeneous MTL setting.

\paragraph{SOTA and direct comparisons.}
The most directly comparable public SOTA is the 7th ABAW MTL leaderboard (same task triad and metric); ABAW8/9 use different task splits and are not directly comparable~\cite{kollias2025advancements,kollias2025emotions}.
The leading entry reaches $P_{MTL}=1.529$ via progressive staged training, task-selective fusion, and temporal context~\cite{liu2024progressive}; the official challenge baseline is $P_{MTL}=0.34$~\cite{kollias20247th}.

\section{Method}
\label{sec:method}

\subsection{Overview}

We propose \emph{AffectFlow-DINO}, a frame-level multi-task model for the ABAW 2026 MTL challenge.
Since s-Aff-Wild2 distributes data as isolated cropped face images rather than video sequences, designing a strictly frame-level system is the faithful choice for this benchmark; temporal sequence reconstruction is treated as an orthogonal future extension rather than a limitation of the current scope.
Given a cropped face image $x$, the model simultaneously predicts continuous valence-arousal (VA), categorical facial expression (EXPR), and binary Action Unit (AU) activations.
Beyond deterministic point prediction, it learns a conditional rectified-flow distribution over the joint affect representation, enabling uncertainty-aware, one-to-many prediction at inference.

The core motivation is the inherent perceptual ambiguity of in-the-wild facial behavior: a subtle smile, partial occlusion, or low-intensity expression may correspond to several plausible VA and AU configurations.
Rather than collapsing this uncertainty to a single point estimate, AffectFlow learns a generative transport map from isotropic Gaussian noise to the conditional target distribution $p(y \mid x)$.
This produces a family of plausible affect vectors per image, whose mean improves over deterministic prediction while their spread reflects prediction uncertainty.

\begin{figure}[t]
  \centering
  \includegraphics[width=\linewidth]{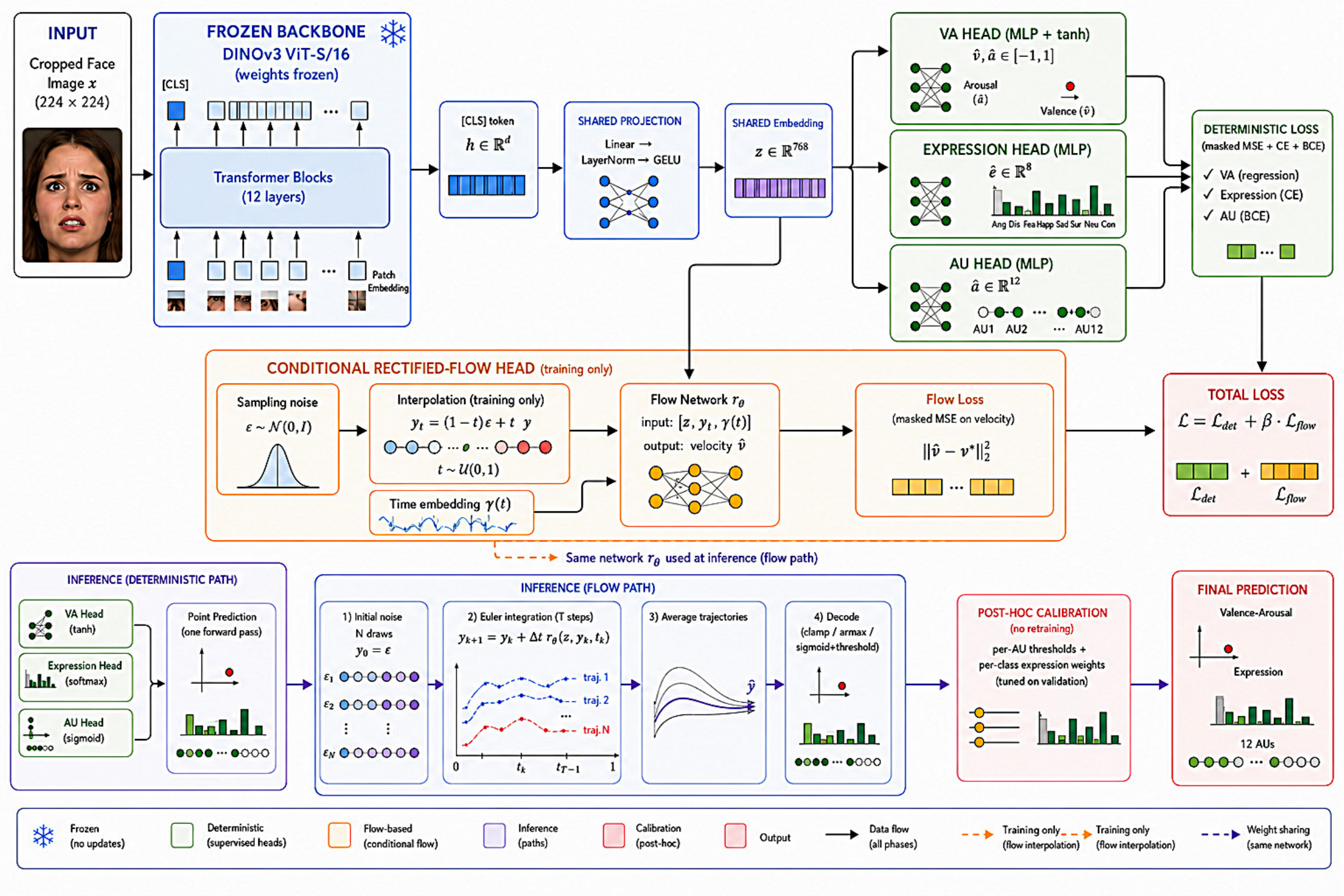}
  \caption{%
    Overview of AffectFlow-DINO.
    A frozen DINOv3 ViT-S/16 backbone encodes the face into a shared embedding $z$.
    Three deterministic task heads produce point predictions for VA, expression, and AUs.
    The conditional rectified-flow head receives $z$ together with a linearly interpolated noised target $y_t$ and a sinusoidal time embedding $\gamma(t)$, and learns to predict the transport velocity.
    At inference, $N$ trajectories are integrated from noise to the target space and averaged to produce the final prediction.
  }
  \label{fig:architecture}
\end{figure}

The model comprises four components, illustrated in Figure~\ref{fig:architecture}:
\begin{enumerate}
  \item a frozen DINOv3 ViT-S/16 backbone that extracts rich facial representations;
  \item a shared projection head mapping backbone features into a compact affect embedding;
  \item three deterministic task heads for VA regression, expression classification, and AU detection;
  \item a conditional rectified-flow head that models the joint affect distribution conditioned on the image embedding.
\end{enumerate}

\subsection{Joint Affect Representation and Masked Supervision}

Each training sample consists of a cropped, aligned face frame $x_i$ paired with up to three partially available annotations.
We represent the full affect target as a 22-dimensional vector:
\begin{equation}
  y_i = \bigl[v_i,\, a_i,\, e_i^{(1)}, \ldots, e_i^{(8)},\, u_i^{(1)}, \ldots, u_i^{(12)}\bigr] \in \mathbb{R}^{22},
  \label{eq:joint_target}
\end{equation}
where $v_i, a_i \in [-1,1]$ are valence and arousal, $e_i \in \{0,1\}^8$ is a one-hot expression vector over eight categories, and $u_i \in \{0,1\}^{12}$ is a binary AU vector.
Embedding heterogeneous targets, continuous VA, categorical expressions, and binary AUs, into a single $\mathbb{R}^{22}$ space is a deliberate design choice that enables a unified Euler integration over a shared continuous manifold (Eq.~\eqref{eq:euler}).
Categorical and binary dimensions are treated as unconstrained real-valued scores during the flow trajectory; discrete predictions are recovered only at the final inference step via argmax and sigmoid thresholding (Eq.~\eqref{eq:average}), avoiding per-task generative processes at the known cost of trajectories visiting off-manifold intermediate values, a trade-off shared with continuous label-space generative models. Step-wise projection is deliberately avoided since rectified flow's straight-line trajectories reliably arrive on the target manifold at $t{=}1$.
The s-Aff-Wild2 annotations are incomplete: a significant fraction of frames are annotated for only one or two tasks.
We therefore construct a per-sample binary validity mask $m_i \in \{0,1\}^{22}$ and apply it to every loss term, ensuring that missing targets neither contribute gradients nor distort the flow training signal.

\subsection{DINOv3 Visual Backbone}

The visual encoder is a DINOv3 ViT-S/16 kept frozen throughout training. Its self-supervised DINO objectives~\cite{oquab2023dinov2} enforce spatial semantic consistency across patch tokens, unlike reconstruction-based pre-training such as MAE~\cite{he2022mae}, making it well-suited to Action Unit detection, where muscle activations manifest as spatially localised deformations.
For each face image $x_i$ resized to $224 \times 224$, the encoder extracts a $d$-dimensional feature vector from the \texttt{[CLS]} token:
\begin{equation}
  h_i = f_{\theta}(x_i) \in \mathbb{R}^{d}.
  \label{eq:backbone}
\end{equation}
A shared projection head then maps $h_i$ to the working embedding dimension:
\begin{equation}
  z_i = \phi\bigl(W_h\, h_i\bigr),
  \label{eq:projection}
\end{equation}
where $\phi$ denotes LayerNorm followed by GELU activation, and $W_h \in \mathbb{R}^{D \times d}$ is a learned linear projection with $D=768$.

\subsection{Deterministic Multi-Task Heads}

Three task-specific two-layer MLP heads with LayerNorm, GELU activation, and dropout are attached to the shared embedding $z_i$:
\begin{align}
  \hat{y}^{VA}_i   &= \tanh\!\bigl(\mathrm{MLP}_{VA}(z_i)\bigr) \in [-1,1]^2, \label{eq:va_head}\\
  \hat{y}^{EXPR}_i &= \mathrm{MLP}_{EXPR}(z_i) \in \mathbb{R}^{8}, \label{eq:expr_head}\\
  \hat{y}^{AU}_i   &= \mathrm{MLP}_{AU}(z_i) \in \mathbb{R}^{12}. \label{eq:au_head}
\end{align}
The $\tanh$ nonlinearity in Eq.~\eqref{eq:va_head} constrains VA predictions to the official annotation range.
Expression logits in Eq.~\eqref{eq:expr_head} are trained with masked cross-entropy, and AU logits in Eq.~\eqref{eq:au_head} with masked binary cross-entropy.
AU activations are obtained at inference by thresholding the sigmoid output.

\subsection{Conditional Rectified Flow}

The deterministic heads yield accurate point predictions but cannot represent the uncertainty and multimodality inherent in ambiguous facial images.
The AffectFlow component addresses this by learning a conditional generative model $p(y \mid x)$ via rectified flow~\cite{liu2022flow}.

\paragraph{Training.}
For each sample $(x_i, y_i)$, we draw noise $\epsilon \sim \mathcal{N}(0, I_{22})$ and a time scalar $t \sim \mathcal{U}(0,1)$, and form the linear interpolant:
\begin{equation}
  y_t = (1 - t)\,\epsilon + t\,y_i.
  \label{eq:interpolant}
\end{equation}
The rectified-flow objective minimizes the squared error between the predicted and the true constant velocity field $v^* = y_i - \epsilon$:
\begin{equation}
  \mathcal{L}_{\text{flow}}
  = \frac{
    \displaystyle\sum_{j=1}^{22} m_i^{(j)}\Bigl(\hat{v}_i^{(j)} - v_i^{*(j)}\Bigr)^2
  }{
    \displaystyle\sum_{j=1}^{22} m_i^{(j)} + \varepsilon
  },
  \label{eq:flow_loss}
\end{equation}
The mask $m_i$ ensures that absent labels do not contribute to the flow training signal.
Formally, zeroing the missing velocity dimensions supervises only the annotated marginal $p(y_{\mathcal{T}} \mid x)$ (where $\mathcal{T}$ denotes the annotated task subset for that frame); under the mild assumption that annotation missingness is independent of true affect values, each masked step is a valid update for the full joint $p(y \mid x)$, and the union of annotated marginals across the dataset collectively constrains the full joint conditional while preserving its marginal consistency.
This assumption is not fully met in practice: if rare expressions or AUs are systematically less likely to be annotated (e.g.\ because annotators skip ambiguous or low-intensity frames disproportionately common to those classes), the annotated marginal is a biased estimate of the true one, and the learned flow could inherit that bias rather than merely inheriting the class imbalance itself.
The velocity $\hat{v}_i$ is predicted by a flow network $r_\psi$:
\begin{equation}
  \hat{v}_i = r_\psi\bigl([z_i,\; y_t,\; \gamma(t)]\bigr),
  \label{eq:flow_network}
\end{equation}
with $\gamma(t)$ a sinusoidal time embedding and $[\cdot]$ denoting concatenation.

\paragraph{Inference.}
To obtain a prediction for image $x$, we draw $N$ independent noise samples $\{\epsilon^{(n)}\}_{n=1}^N$, each initialized at $y_0^{(n)} = \epsilon^{(n)}$.
Each sample is propagated forward with $T$ Euler integration steps of step size $\Delta t = 1/T$:
\begin{equation}
  y_{k+1}^{(n)} = y_k^{(n)} + \Delta t\; r_\psi\!\bigl([z,\; y_k^{(n)},\; \gamma(t_k)]\bigr), \quad k = 0, \ldots, T-1,
  \label{eq:euler}
\end{equation}
where $t_k = k\,\Delta t$.
The final predictions are obtained by averaging over the $N$ trajectories at $t=1$:
\begin{align}
  \bar{y} &= \tfrac{1}{N}\sum_{n=1}^N y_T^{(n)}, \label{eq:average}\\
  \hat{y}^{VA}   &= \text{clamp}\bigl(\bar{y}_{1:2},\; -1, 1\bigr), \nonumber\\
  \hat{y}^{EXPR} &= \arg\max\bigl(\bar{y}_{3:10}\bigr), \nonumber\\
  \hat{y}^{AU}_k &= \mathbf{1}\bigl[\sigma(\bar{y}_{10+k}) > \tau\bigr], \quad k=1,\ldots,12. \nonumber
\end{align}
The threshold $\tau$ is a tunable decoding parameter; we ablate it across validation data in Section~\ref{sec:experiments}.
This inference procedure contrasts fundamentally with deterministic MTL: the same face image generates a \emph{distribution} over affect states, providing uncertainty estimates alongside the official prediction.

\subsection{Training Objective}

The full training objective is a weighted combination of the deterministic multi-task loss and the flow loss:
\begin{equation}
  \mathcal{L}
  = \mathcal{L}_{\text{det}} + \beta\,\mathcal{L}_{\text{flow}},
  \label{eq:total_loss}
\end{equation}
where
\begin{equation}
  \mathcal{L}_{\text{det}}
  = \lambda_{VA}\,\mathcal{L}_{VA}
  + \lambda_{EXPR}\,\mathcal{L}_{EXPR}
  + \lambda_{AU}\,\mathcal{L}_{AU}.
  \label{eq:det_loss}
\end{equation}
The flow weight $\beta$ controls the balance between learning a high-quality generative model and optimizing task metrics directly.
An ablation over $\beta \in \{0.25, 0.5, 1.0, 2.0\}$ (Appendix~\ref{sec:app_e4}) finds that $\beta \in \{0.5, 1.0\}$ achieve equivalent best validation $P_{MTL}$; we adopt $\beta = 1.0$ as the default.
The model is optimized with AdamW and the best checkpoint is selected by validation $P_{MTL} = P_{VA} + P_{EXPR} + P_{AU}$.

\section{Experiments}
\label{sec:experiments}

\subsection{Experimental Setup}

All models are trained on the official ABAW 2026 frame-level training split of s-Aff-Wild2 and evaluated on the provided validation split.
The training set contains 142,382 frames with partially available labels: 103,917 valid VA, 90,645 valid expression, and 103,316 valid AU rows.
The validation set contains 26,876 frames with full VA and AU labels and 15,440 valid expression labels.
All invalid labels are excluded via per-task binary masks as described in Section~\ref{sec:method}.

\noindent\textit{Label statistics and class imbalance.}
Expression labels exhibit an 8$\times$ imbalance: Other (27.4\%), Neutral (26.5\%), and Happiness (20.0\%) dominate, while Fear (3.4\%) and Disgust (3.5\%) together account for under 7\% of valid labels.
For AUs, AU25 is active in 68\% of valid training rows while AU15, AU23, and AU24 are positive in only 2--5\%, yielding a 29$\times$ overall imbalance; full per-class statistics are in Appendix~\ref{app:datastats} (Table~\ref{tab:dataset_stats}).
Notably, joint three-task supervision is available for only 36.6\% of training frames, making per-task masking essential.

Unless stated otherwise, all experiments share the following configuration: a frozen DINOv3 ViT-S/16 backbone, input resolution $224\times224$, AdamW optimizer (lr=1e-4, weight decay=1e-2), 20 training epochs, batch size 64, and checkpoint selection by the validation composite metric $P_{MTL} = P_{VA} + P_{EXPR} + P_{AU}$, where $P_{VA}$ is the mean CCC of valence and arousal, $P_{EXPR}$ is expression macro-F1, and $P_{AU}$ is mean AU F1.
For flow-based evaluation, predictions are produced with $N=16$ samples and $T=30$ Euler steps unless the ablation varies those parameters.

\subsection{What Does Each Training Objective Contribute?}
\label{sec:e1}

We isolate the contribution of each training objective by comparing three training regimes: deterministic supervision only ($\beta=0$, no flow loss); flow supervision only (no deterministic task losses, $\lambda_{det}=0$); and both objectives jointly ($\beta=1$), which we call \emph{AffectFlow}.
Each checkpoint is decoded both deterministically and with flow sampling, yielding the six-condition study in Table~\ref{tab:e1_e3_comparison}.

\begin{table}[t]
  \caption{%
    Contribution of each training objective.
    Det-only: $\beta=0$; Flow-only: $\lambda_{det}=0$, $\beta=1$; AffectFlow: $\beta=1$, both objectives jointly.
    Rows marked ``sanity'' verify that applying a decode mode the model was not trained for is harmful.
  }
  \label{tab:e1_e3_comparison}
  \centering
  \begin{tabular}{@{}llrrrrrr@{}}
    \toprule
    Model & Decode & CCC-V & CCC-A & $P_{VA}$ & $P_{EXPR}$ & $P_{AU}$ & $P_{MTL}$ \\
    \midrule
    Official baseline & -- & -- & -- & -- & -- & -- & 0.4500 \\
    Det-only ($\beta=0$) & Deterministic    & 0.234 & 0.163 & 0.199 & 0.216 & 0.378 & 0.793 \\
    Det-only ($\beta=0$) & Flow (sanity)    & $-$0.017 & $-$0.003 & $-$0.010 & 0.127 & 0.285 & 0.402 \\
    Flow-only ($\lambda_{det}=0$) & Flow   & 0.253 & 0.169 & 0.211 & 0.191 & 0.371 & 0.773 \\
    Flow-only ($\lambda_{det}=0$) & Det (sanity) & 0.062 & $-$0.003 & 0.030 & 0.117 & 0.262 & 0.408 \\
    AffectFlow ($\beta=1$) & Deterministic & 0.232 & 0.168 & 0.200 & 0.218 & 0.384 & 0.802 \\
    AffectFlow ($\beta=1$) & Flow          & 0.290 & 0.186 & 0.238 & 0.212 & 0.376 & \textbf{0.826} \\
    \bottomrule
  \end{tabular}
\end{table}

The det-only sanity check (det-only checkpoint decoded with flow) collapses to $P_{MTL}=0.402$, and the flow-only sanity check (flow-only checkpoint decoded deterministically) collapses to $P_{MTL}=0.408$: each decode mode requires its corresponding training objective.
Flow-only training with flow decoding reaches $P_{MTL}=0.773$, confirming that the flow head does learn a meaningful conditional distribution even without deterministic task supervision.
However, AffectFlow's flow decoding ($P_{MTL}=0.826$) substantially outperforms flow-only training ($P_{MTL}=0.773$), revealing that deterministic supervision acts as an auxiliary training signal for the flow head: the task losses encourage the shared embedding to encode affect-discriminative features, which in turn improve the quality of the learned flow trajectories.
Symmetrically, deterministic heads trained without the flow objective (det-only) perform comparably to those in AffectFlow under deterministic decoding (0.793 vs.\ 0.802), indicating that the flow loss does not harm the task heads but adds the generative pathway as an orthogonal capability.
Flow decoding on trained AffectFlow yields large gains in CCC-V ($+0.058$) and CCC-A ($+0.018$) relative to its own deterministic decode, lifting $P_{VA}$ from 0.200 to 0.238 at a modest cost to $P_{AU}$.
This pattern, flow sampling improving VA most, is consistent throughout all ablations: continuous targets benefit most from distributional averaging, while categorical and binary tasks benefit somewhat less.

Flow loss weight $\beta$ and inference efficiency sweeps are in Appendix~\ref{sec:app_e4_e5e6}: $\beta \in [0.5, 1.0]$ is a stable plateau (we adopt $\beta=1.0$); flow sampling saturates at $N=8$ samples and $T=10$ Euler steps (we use $N=16$, $T=30$), consistent with the straight-line trajectory property of rectified flow~\cite{liu2022flow}.
Minor sweeps for VA loss weight and global AU threshold yield no improvement (Appendix~\ref{sec:e7}, \ref{sec:e8}); the latter motivates per-AU calibration in Section~\ref{sec:e14}.

\subsection{Low-Learning-Rate DINOv3 Fine-Tuning}
\label{sec:e9}

The frozen-backbone experiments isolate the contribution of AffectFlow, but they leave the visual representation unchanged.
We therefore fine-tune the DINOv3 backbone end-to-end with a smaller learning rate ($10^{-5}$), batch size 32, and the same 20-epoch schedule.
Table~\ref{tab:e9_finetune} compares deterministic and flow decoding for the fine-tuned checkpoint.

\begin{table}[t]
  \caption{Effect of low-learning-rate DINOv3 fine-tuning. Both rows use the same fine-tuned checkpoint.}
  \label{tab:e9_finetune}
  \centering
  \begin{tabular}{@{}lrrrrrr@{}}
    \toprule
    Decode & CCC-V & CCC-A & $P_{VA}$ & $P_{EXPR}$ & $P_{AU}$ & $P_{MTL}$ \\
    \midrule
    Flow          & 0.336 & 0.279 & 0.307 & 0.236 & 0.385 & 0.928 \\
    Deterministic & 0.352 & 0.285 & 0.318 & 0.285 & 0.441 & \textbf{1.045} \\
    \bottomrule
  \end{tabular}
\end{table}

Fine-tuning provides the largest gain observed so far, raising $P_{MTL}$ from the best frozen-backbone result of 0.831 to 1.045.
Interestingly, deterministic decoding outperforms flow decoding after fine-tuning.
This suggests that once the visual encoder is adapted to the target domain, the deterministic heads become substantially stronger, while the flow head may require retuning of its loss weight, sampling temperature, or training schedule.

We explored four expression-specific modifications to address the majority-class collapse: expression loss weight scaling, focal loss, per-task separate projection heads, and their combination.
All four fail to improve $P_{EXPR}$ over the AffectFlow baseline and consistently degrade $P_{VA}$; full results are in Appendix~\ref{sec:e10}.
The common failure mode is that scaling or reweighting the cross-entropy loss, regardless of form, cannot overcome the gradient domination by Neutral and Other without structural sampling changes.
This motivates the class-balanced sampling and positive-weighting strategies evaluated below.

We also explored \emph{patch soft-pool} (PSP) aggregation, which learns per-channel attention weights over the 196 frozen ViT patch tokens and concatenates the result with the CLS token (Appendix~\ref{sec:app_psp}).
PSP deterministic reaches $P_{MTL}=0.859$, the best frozen-backbone result without calibration, by capturing spatially-localised AU information, but is not combined with backbone fine-tuning in this study.

\subsection{Per-AU Threshold Calibration}
\label{sec:e14}

Appendix~\ref{sec:e8} showed that a single global AU threshold cannot improve over $\tau=0.5$.
We therefore tune an independent threshold per AU on the validation set by grid search, leaving VA and expression predictions unchanged.
Table~\ref{tab:e14_perau} reports the gain from this post-hoc step on three representative checkpoints.

\begin{table}[t]
  \caption{Per-AU threshold calibration (post-hoc). VA and $P_{EXPR}$ are unchanged; only $P_{AU}$ and $P_{MTL}$ improve.}
  \label{tab:e14_perau}
  \centering\small
  \setlength{\tabcolsep}{3pt}
  \begin{tabular}{@{}lrrrr@{}}
    \toprule
    Checkpoint & $P_{AU}$ (def.) & $P_{AU}$ (cal.) & $P_{MTL}$ (def.) & $P_{MTL}$ (cal.) \\
    \midrule
    AffectFlow, Flow ($N$=16, $T$=30) & 0.376 & 0.441 & 0.823 & \textbf{0.888} \\
    Separate projection heads, Flow    & 0.371 & 0.436 & 0.797 & 0.862 \\
    Fine-tuned backbone, Det                  & 0.441 & 0.497 & 1.045 & \textbf{1.101} \\
    \bottomrule
  \end{tabular}
\end{table}

Per-AU calibration is theoretically principled rather than an ad-hoc leaderboard trick: the AU head predicts sigmoid activation probabilities $p(AU_k{=}1 \mid x)$ conditioned on the image, so the optimal decision boundary need not equal 0.5, particularly when class-imbalanced BCE training systematically suppresses predicted probabilities for rare AUs.
This suppression is confirmed by the per-class analysis in Section~\ref{sec:e15}: AU15 and AU23 have near-zero F1 at threshold 0.5 despite non-trivial true positive rates, meaning the model has learned genuine discriminative signal that the default threshold cannot recover.
The rectified-flow head further reinforces this framing: by modeling $p(y \mid x)$ as a full conditional distribution, AffectFlow provides per-sample uncertainty information alongside point predictions; per-AU calibration exploits this probabilistic output to find the threshold at which the flow-informed sigmoid posterior best separates positives from negatives.
Per-AU calibration yields a consistent $P_{AU}$ gain of approximately +0.06 across all checkpoints, with no retraining cost.
Rare AUs such as AU15 and AU23 benefit most from lower thresholds, while frequently active AUs (AU25, AU7) prefer thresholds near 0.55--0.60.
The best overall validation score is $P_{MTL}=1.101$, obtained by combining low-LR DINOv3 fine-tuning with deterministic decoding and per-AU calibration.
An analogous post-hoc calibration strategy applied to expression class predictions (Section~\ref{sec:e27}) yields a further $+0.054$ $P_{MTL}$ gain on top of per-AU calibration.

\subsection{Per-Class Analysis and Class-Weighted Expression Loss}
\label{sec:e15}

The macro-F1 metric used for checkpoint selection conceals large per-class disparities.
Full per-class expression F1 for the AffectFlow and fine-tuned models is in Appendix~\ref{app:perau} (Table~\ref{tab:e15_perclass}), alongside the per-AU breakdown (Table~\ref{tab:e15_perau}); Happiness and Other achieve $F1 > 0.49$, but Fear ($F1 = 6.3$--$6.9\%$) and Sadness ($F1 = 4.3$--$7.1\%$) are catastrophically low even after fine-tuning.
The model collapses 77\% of Sadness, 64\% of Surprise, and 41\% of Fear into ``Other'', which absorbs the bulk of gradient mass under standard cross-entropy.
AU prediction mirrors the same pattern: AU15 and AU23 have near-zero F1 at $\tau=0.5$ despite non-trivial true positive rates (see Appendix~\ref{app:perau}).

\paragraph{Class-weighted cross-entropy.}
We replace standard cross-entropy with an inverse-frequency weighted variant assigning each class $c$ weight $w_c = N_\text{valid}/(8\,n_c)$ (Appendix~\ref{app:datastats}, Table~\ref{tab:dataset_stats}).
After seven training epochs (deterministic validation, frozen backbone), the best checkpoint reaches $P_{EXPR} = 0.239$ and $P_{MTL} = 0.834$, compared to $P_{EXPR} = 0.218$ and $P_{MTL} = 0.802$ for the AffectFlow baseline at the same evaluation protocol.
The improvement is concentrated in the minority classes: Disgust F1 rises from $11.8\%$ to $39.0\%$ and Surprise from $2.8\%$ to $12.8\%$, while Fear ($0.3\%$) and Sadness ($8.5\%$) remain the primary bottleneck, suggesting that their imbalance severity requires additional intervention beyond loss reweighting.

Imbalance remedies are detailed in Appendix~\ref{app:imbalance}.
In brief: a class-balanced \texttt{WeightedRandomSampler} raises $P_{EXPR}$ to $0.251$ ($+0.039$ over AffectFlow flow), the best frozen-backbone expression result; per-AU BCE positive weighting raises $P_{AU}$ to $0.427$--$0.439$, the best frozen-backbone AU result without calibration; and label smoothing (Appendix~\ref{sec:e18}) and global AU loss weight (Appendix~\ref{sec:e21}) provide no meaningful gains.
Critically, AU positive-weighted training is incompatible with flow decoding ($P_{MTL}$ degrades to $0.745$--$0.777$) because the reweighted loss shifts predicted probability distributions incompatibly with the rectified-flow sampler.

\subsection{Flow Retuning after Backbone Fine-Tuning}
\label{sec:e22}

Section~\ref{sec:e9} showed that low-LR fine-tuning of the DINOv3 backbone improves all tasks substantially, but leaves the flow head undertuned: deterministic decoding outperforms flow decoding in the fine-tuned setting.
We address this by reactivating the flow objective ($\beta=0.5$) during fine-tuning, retuning the flow head jointly with the backbone.
Table~\ref{tab:e22_flowretune} compares fine-tuning with a frozen flow head against fine-tuning with flow retuning under both decode modes.

\begin{table}[t]
  \caption{Flow retuning after backbone fine-tuning for $\beta \in \{0.5, 1.0\}$. $\dagger$~per-AU threshold calibration applied.}
  \label{tab:e22_flowretune}
  \centering\small
  \setlength{\tabcolsep}{3pt}
  \begin{tabular}{@{}llrrrrrr@{}}
    \toprule
    Config & Decode & CCC-V & CCC-A & $P_{VA}$ & $P_{EXPR}$ & $P_{AU}$ & $P_{MTL}$ \\
    \midrule
    Fine-tuned              & Det                    & 0.352 & 0.285 & 0.318 & 0.285 & 0.441 & 1.045 \\
    Fine-tuned$^\dagger$    & Det                    & 0.352 & 0.285 & 0.318 & 0.285 & 0.497 & 1.101 \\
    Fine-tuned+flow retune ($\beta{=}0.5$) & Flow ($N$=16, $T$=30) & 0.296 & 0.284 & 0.290 & 0.279 & 0.386 & 0.956 \\
    Fine-tuned+flow retune ($\beta{=}0.5$) & Det                   & 0.308 & 0.273 & 0.291 & 0.303 & 0.469 & 1.062 \\
    \midrule
    Fine-tuned+flow retune ($\beta{=}1.0$) & Flow ($N$=16, $T$=30) & 0.313 & 0.281 & 0.297 & 0.252 & 0.382 & 0.931 \\
    Fine-tuned+flow retune ($\beta{=}1.0$) & Det                   & 0.387 & 0.263 & 0.325 & 0.296 & 0.453 & 1.073 \\
    Fine-tuned+flow retune ($\beta{=}1.0$)$^\dagger$ & Det         & 0.387 & 0.263 & 0.325 & 0.296 & 0.502 & \textbf{1.123} \\
    \bottomrule
  \end{tabular}
\end{table}

Flow retuning at $\beta{=}0.5$ under deterministic decoding reaches $P_{MTL}=1.062$, surpassing plain fine-tuning without calibration ($1.045$), while flow decoding ($P_{MTL}=0.956$) substantially outperforms the AffectFlow frozen-backbone flow baseline ($0.826$), confirming that the flow head recovers distributional capability after backbone adaptation.

Increasing the flow weight to $\beta{=}1.0$ further improves deterministic decoding to $P_{MTL}=1.073$, driven by a large CCC-V gain ($0.387$ vs.\ $0.308$ for $\beta{=}0.5$) at a modest cost to CCC-A and $P_{AU}$.
After per-AU calibration, this configuration reaches $P_{MTL}=1.123$, surpassing the calibrated plain fine-tuning baseline of $1.101$; post-hoc expression calibration (Section~\ref{sec:e27}) further raises this to $P_{MTL}=1.177$.
Flow decoding for $\beta{=}1.0$ ($P_{MTL}=0.931$) trails the $\beta{=}0.5$ flow variant ($0.956$), suggesting that the stronger flow objective reshapes the learned target distribution in a way that marginally reduces flow-sampled consistency while benefiting the deterministic readout.
In both variants, deterministic decoding outperforms flow decoding after retuning, indicating that further alignment of the flow head with the adapted backbone remains an open direction.

Kitchen-sink fine-tuning (Appendix~\ref{app:e23}) applies class-weighted cross-entropy and per-AU positive weighting jointly during backbone fine-tuning.
The combination does not improve $P_{MTL}$ over plain fine-tuning: $P_{MTL}=1.041$ vs.\ $1.045$, with the AU gain ($+0.042$) offset by a $P_{VA}$ drop ($-0.047$).
This confirms that frozen-backbone reweighting strategies do not stack with backbone adaptation: once the backbone is fine-tuned, flow retuning is the more effective path.
Replacing ViT-S with ViT-B improves $P_{VA}$ ($0.354$ vs.\ $0.325$) but degrades $P_{EXPR}$ ($0.255$ vs.\ $0.296$), yielding $P_{MTL}=1.116$ calibrated; horizontal-flip test-time augmentation reduces all scores to $P_{MTL}=1.111$ calibrated, full results in Appendix~\ref{sec:app_e24_e26}.

\subsection{Per-Class Expression Threshold Calibration}
\label{sec:e27}

Section~\ref{sec:e14} showed that per-AU threshold calibration yields a consistent $+0.06$ $P_{AU}$ gain by finding per-AU decision boundaries that maximise F1 on the validation set.
We apply the same post-hoc principle to the expression head: per-class logit weights are tuned on the validation set to maximise macro-F1, directly addressing the systematic under-prediction of minority classes (Fear, Sadness) caused by the $8\times$ expression imbalance.
The full per-class breakdown is in Appendix~\ref{app:perau} (Table~\ref{tab:e27_exprcal}, Figure~\ref{fig:expr_calibration}).
Fear benefits most ($0.038 \to 0.331$, $+0.293$), followed by Sadness ($0.171 \to 0.282$, $+0.111$) and Disgust ($0.465 \to 0.507$, $+0.042$); majority classes (Happiness, Other) are essentially unchanged.
The large Fear gain confirms that the model has learned genuine discriminative signal for the rarest class: the default argmax simply never predicts it because its logit magnitude is suppressed relative to Neutral and Other by the imbalanced training distribution.
Calibration recovers this latent signal without any retraining cost, in direct analogy to how per-AU threshold calibration recovers rare-AU signal (AU15, AU23) in Section~\ref{sec:e14}.

Combining per-class expression calibration with per-AU calibration on the fine-tuned+flow retune ($\beta{=}1.0$) checkpoint yields the best overall result:
$P_{VA}=0.325$ $+$ $P_{EXPR}=0.350$ $+$ $P_{AU}=0.502$ $=$ $\mathbf{P_{MTL}=1.177}$, a $+0.054$ improvement over the flow-retuned, per-AU-calibrated baseline ($1.123$).

\subsection{Summary of Ablation Results}
\label{sec:results_summary}

Table~\ref{tab:results_summary} collects the best representative result per configuration (all models use frozen DINOv3 unless noted; full 18-configuration table in Appendix~\ref{app:full_results}).
Both training objectives are needed: flow-only ($P_{MTL}=0.773$) and det-only-with-flow-decode ($0.402$) both underperform jointly-trained AffectFlow ($0.826$), and patch soft-pool further lifts the frozen-backbone deterministic result to $0.859$.
The best frozen-backbone result without calibration is $0.890$ (AU positive weighting combined with class-weighted cross-entropy; Appendix~\ref{app:imbalance}), while per-AU calibration alone raises AffectFlow flow to $0.888$.
Backbone fine-tuning is the dominant lever, lifting the best result to $1.045$ ($1.101$ calibrated); flow retuning ($\beta{=}1.0$) on top of it reaches $1.073$ ($1.123$ calibrated), and post-hoc expression calibration adds a further $+0.054$ for the best overall result $\mathbf{P_{MTL}=1.177}$.
Kitchen-sink fine-tuning (Appendix~\ref{app:e23}) and all expression/AU reweighting ablations (Appendices~\ref{app:ablations}, \ref{app:imbalance}) do not compound with backbone fine-tuning.

\begin{table}[t]
  \caption{%
    Best validation result per configuration (frozen backbone unless noted); abridged, full table (18 configurations) in Appendix~\ref{app:full_results}.
    \emph{AF} = AffectFlow; \emph{Det} = deterministic decode; \emph{Flow} = flow decode ($N$=16, $T$=30 unless varied);
    $\beta$ = flow loss weight; \emph{FT} = low-LR backbone fine-tuning.
  }
  \label{tab:results_summary}
  \centering\small
  \setlength{\tabcolsep}{4pt}
  \renewcommand{\arraystretch}{0.88}
  \begin{tabular}{@{}lcccccc@{}}
    \toprule
    Configuration & CCC-V & CCC-A & $P_{VA}$ & $P_{EXPR}$ & $P_{AU}$ & $P_{MTL}$ \\
    \midrule
    Official baseline                         & --    & --    & --    & --    & --    & 0.450 \\
    DINOv3 Det, no flow ($\beta{=}0$)        & 0.234 & 0.163 & 0.199 & 0.216 & 0.378 & 0.793 \\
    AF $\beta{=}0.5$, Flow, $T{=}30$         & 0.289 & 0.188 & 0.239 & 0.211 & 0.376 & 0.826 \\
    AF $\beta{=}1$, Flow + per-AU cal        & 0.288 & 0.185 & 0.237 & 0.210 & 0.441 & 0.888 \\
    PSP CLS+patch, Det                       & 0.271 & 0.198 & 0.235 & 0.236 & 0.389 & 0.859 \\
    AU pos-wt + CW-CE, Det                    & 0.259 & 0.163 & 0.211 & 0.240 & 0.439 & 0.890 \\
    AF FT, Det                                & 0.352 & 0.285 & 0.318 & 0.285 & 0.441 & 1.045 \\
    AF FT, Det + per-AU cal                   & 0.352 & 0.285 & 0.318 & 0.285 & 0.497 & 1.101 \\
    FT+Flow retune $\beta{=}1.0$, Det        & 0.387 & 0.263 & 0.325 & 0.296 & 0.453 & 1.073 \\
    FT+Flow retune $\beta{=}1.0$, Det+per-AU cal & 0.387 & 0.263 & 0.325 & 0.296 & 0.502 & 1.123 \\
    ViT-B FT, Det+per-AU cal                 & 0.416 & 0.293 & 0.354 & 0.255 & 0.507 & 1.116 \\
    FT+Flow retune, AU cal+expr cal          & 0.387 & 0.263 & 0.325 & 0.350 & 0.502 & \textbf{1.177} \\
    \bottomrule
  \end{tabular}
\end{table}

\subsection{Discussion and Open Challenges}

All results in this paper are on the local s-Aff-Wild2 validation split; the SOTA scores cited in Section~\ref{sec:related_work} are from the 7th ABAW hidden test set and are not directly comparable.
The current best validated result ($P_{MTL}=1.177$) substantially improves over the official baseline ($P_{MTL}=0.45$); Figure~\ref{fig:ablation_progression} in Appendix~\ref{app:ablations} visualises the full progression across key configurations.
Within our controlled ablation study the gap is most pronounced in $P_{EXPR}$ (best uncalibrated: $0.296$; best calibrated: $0.350$); two principal factors account for it.
First, expression prediction remains the most challenging task: the per-class breakdown (Table~\ref{tab:e15_perclass}) reveals that Fear and Sadness F1 remain below 7\% at the default threshold, with predictions collapsed into ``Other''.
The $8\times$ expression imbalance is the proximate cause; class-weighted cross-entropy and balanced sampling partially close the gap ($P_{EXPR}=0.239$--$0.251$), while post-hoc expression calibration recovers the latent rare-class signal directly, raising Fear F1 from $0.038$ to $0.331$ and Sadness from $0.171$ to $0.282$ without retraining.
Second, the flow head requires retuning after backbone fine-tuning to recover distributional prediction capability: reactivating the flow objective during fine-tuning is beneficial, with $\beta{=}1.0$ yielding the strongest deterministic result ($P_{MTL}=1.073$; $1.123$ with per-AU calibration; $1.177$ with combined calibration), though deterministic decoding continues to outperform flow decoding in all fine-tuned variants.
This suggests the flow objective acts as a \emph{structure-aware regularizer} for $z$, enforcing joint affect structure that benefits the deterministic heads even when flow decoding is not used at inference.

\section{Conclusion}
\label{sec:conclusion}

We introduced AffectFlow-DINO, a multi-task facial affect model that treats in-the-wild facial behavior as a conditional distribution $p(y \mid x)$ rather than a deterministic mapping, learned via a rectified-flow head over the joint 22-dimensional affect space and validated through 26 controlled ablations on s-Aff-Wild2.

Three lessons emerge.
The flow and deterministic objectives are complementary: flow-only training collapses without the deterministic anchor, and deterministic-only inference discards distributional capability.
Flow sampling benefits VA estimation most, the task most sensitive to continuous ambiguity, and saturates rapidly, consistent with rectified flow's straight-line transport.
Backbone adaptation is the dominant lever, but the flow head must be retrained jointly during fine-tuning, or the generative pathway degrades.

A broader implication is that severe class imbalance suppresses learned signals without erasing them: post-hoc calibration unlocks near-zero F1 for rare categories (Fear, Sadness, AU15, AU23) at zero cost, suggesting decision boundaries can surface rare-class signal already encoded in the representation.

Closing the gap to SOTA likely requires temporal aggregation, more expressive fine-tuned decoding, and joint backbone-flow multi-task scaling from the start.

\bibliographystyle{splncs04}
\bibliography{main}

\appendix

\section{Full Results Summary}
\label{app:full_results}

Table~\ref{tab:results_summary_full} is the unabridged version of the main paper's results summary (Section~\ref{sec:results_summary}), including intermediate and negative-result configurations omitted there for space.

\begin{table}[t]
  \caption{%
    Best validation result per configuration (frozen backbone unless noted); unabridged version of Table~\ref{tab:results_summary}.
    \emph{AF} = AffectFlow; \emph{Det} = deterministic decode; \emph{Flow} = flow decode ($N$=16, $T$=30 unless varied);
    $\beta$ = flow loss weight; \emph{FT} = low-LR backbone fine-tuning.
  }
  \label{tab:results_summary_full}
  \centering\small
  \setlength{\tabcolsep}{4pt}
  \begin{tabular}{@{}lcccccc@{}}
    \toprule
    Configuration & CCC-V & CCC-A & $P_{VA}$ & $P_{EXPR}$ & $P_{AU}$ & $P_{MTL}$ \\
    \midrule
    Official baseline                         & --    & --    & --    & --    & --    & 0.450 \\
    DINOv3 Det, no flow ($\beta{=}0$)        & 0.234 & 0.163 & 0.199 & 0.216 & 0.378 & 0.793 \\
    AF $\beta{=}0.5$, Flow, $T{=}30$         & 0.289 & 0.188 & 0.239 & 0.211 & 0.376 & 0.826 \\
    AF $\beta{=}1$, Flow, $T{=}50$           & 0.287 & 0.188 & 0.237 & 0.219 & 0.375 & 0.831 \\
    AF $\beta{=}1$, Flow + per-AU cal        & 0.288 & 0.185 & 0.237 & 0.210 & 0.441 & 0.888 \\
    PSP CLS+patch, Flow                      & 0.282 & 0.139 & 0.210 & 0.210 & 0.381 & 0.801 \\
    PSP CLS+patch, Det                       & 0.271 & 0.198 & 0.235 & 0.236 & 0.389 & 0.859 \\
    Balanced sampler, Flow                    & 0.264 & 0.213 & 0.238 & 0.251 & 0.375 & 0.864 \\
    AU pos-wt, Det                            & 0.261 & 0.167 & 0.214 & 0.237 & 0.427 & 0.878 \\
    AU pos-wt + CW-CE, Det                    & 0.259 & 0.163 & 0.211 & 0.240 & 0.439 & 0.890 \\
    AF FT, Det                                & 0.352 & 0.285 & 0.318 & 0.285 & 0.441 & 1.045 \\
    AF FT, Det + per-AU cal                   & 0.352 & 0.285 & 0.318 & 0.285 & 0.497 & 1.101 \\
    FT+Flow retune $\beta{=}0.5$, Det        & 0.308 & 0.273 & 0.291 & 0.303 & 0.469 & 1.062 \\
    Kitchen-sink FT+CW+pos-wt, Det            & 0.319 & 0.223 & 0.271 & 0.288 & 0.483 & 1.041 \\
    FT+Flow retune $\beta{=}1.0$, Det        & 0.387 & 0.263 & 0.325 & 0.296 & 0.453 & 1.073 \\
    FT+Flow retune $\beta{=}1.0$, Det+per-AU cal & 0.387 & 0.263 & 0.325 & 0.296 & 0.502 & 1.123 \\
    ViT-B FT, Det+per-AU cal                 & 0.416 & 0.293 & 0.354 & 0.255 & 0.507 & 1.116 \\
    FT+Flow retune, AU cal+expr cal          & 0.387 & 0.263 & 0.325 & 0.350 & 0.502 & \textbf{1.177} \\
    \bottomrule
  \end{tabular}
\end{table}

\section{Related Datasets}
\label{app:datasets}

Table~\ref{tab:related_datasets} provides an overview of datasets relevant to ABAW-style joint affect modeling.
s-Aff-Wild2 is the only benchmark that combines all three task targets (VA, EXPR, AU) as static image crops, making it the natural fit for this challenge.

\begin{table}[t]
  \caption{Datasets related to ABAW-style affective behavior analysis. ``All 3'' indicates joint availability of valence-arousal, expression, and AU labels in the same benchmark family.}
  \label{tab:related_datasets}
  \centering
  \scriptsize
  \resizebox{\linewidth}{!}{%
  \begin{tabular}{@{}llllp{5.6cm}@{}}
    \toprule
    Dataset & Modality & Setting & Main labels & Relevance to ABAW MTL \\
    \midrule
    Aff-Wild / Aff-Wild2 & Video / A/V & In-the-wild & VA, EXPR, AU & Core ABAW benchmark family; Aff-Wild2 is the main source for joint affect supervision. \\
    s-Aff-Wild2 & Images & In-the-wild & VA, EXPR, AU & Static Aff-Wild2 subset used by recent ABAW MTL challenges, including the local 11th ABAW files. \\
    AffectNet & Images & In-the-wild & EXPR, VA & Useful for large-scale expression and VA pretraining, but not a full AU benchmark. \\
    RAF-DB & Images & In-the-wild & EXPR & Common categorical expression pretraining/evaluation dataset. \\
    DFEW / FERV39k & Video & In-the-wild & EXPR & Dynamic FER datasets useful for temporal facial expression representation learning. \\
    EmotioNet & Images & In-the-wild & AU, EXPR & AU-oriented large-scale facial behavior data; useful for AU pretraining. \\
    BP4D / DISFA & Video & Controlled & AU & Strong AU references, but less representative of unconstrained ABAW conditions. \\
    AFEW-VA / SEWA / OMG-Emotion & Video or A/V & In-the-wild & VA & Related continuous affect datasets, especially for temporal and audiovisual VA modeling. \\
    C-EXPR-DB / BAH & Video or A/V & In-the-wild & Compound expressions, AH & Recent ABAW extensions beyond the classic MTL triad. \\
    \bottomrule
  \end{tabular}}
\end{table}

\section{Flow Weight, Inference Efficiency, and Patch Soft-Pool}
\label{sec:app_e4_e5e6}

\subsection{Flow Loss Weight}
\label{sec:app_e4}

Table~\ref{tab:e4_flowweight} sweeps the flow loss weight $\beta \in \{0.25, 0.5, 1.0, 2.0\}$, with all other hyperparameters fixed and flow decoding at inference ($N=16$, $T=30$).

\begin{table}[t]
  \caption{Effect of flow loss weight $\beta$. Flow decoding with $N=16$, $T=30$.}
  \label{tab:e4_flowweight}
  \centering\small
  \setlength{\tabcolsep}{4pt}
  \begin{tabular}{@{}lrrrrrr@{}}
    \toprule
    $\beta$ & CCC-V & CCC-A & $P_{VA}$ & $P_{EXPR}$ & $P_{AU}$ & $P_{MTL}$ \\
    \midrule
    0.25 & 0.280 & 0.183 & 0.231 & 0.214 & 0.376 & 0.821 \\
    0.5  & 0.289 & 0.188 & 0.239 & 0.211 & 0.376 & \textbf{0.826} \\
    1.0  & 0.290 & 0.186 & 0.238 & 0.212 & 0.376 & 0.826 \\
    2.0  & 0.253 & 0.199 & 0.226 & 0.193 & 0.379 & 0.798 \\
    \bottomrule
  \end{tabular}
\end{table}

Performance is stable for $\beta \in [0.5, 1.0]$; at $\beta=2.0$ the flow objective dominates and $P_{MTL}$ drops.
We adopt $\beta=1.0$ as the default.

\subsection{Inference Efficiency}
\label{sec:app_e5e6}

Table~\ref{tab:e5_e6} sweeps sample count $N$ (left) and integration steps $T$ (right) on the AffectFlow checkpoint.

\begin{table}[t]
  \caption{Inference efficiency on the AffectFlow checkpoint. \emph{Left}: sample count ($T=30$). \emph{Right}: Euler steps ($N=16$).}
  \label{tab:e5_e6}
  \centering\small
  \setlength{\tabcolsep}{4pt}
  \begin{minipage}[t]{0.47\linewidth}
    \centering
    \begin{tabular}{@{}lrrr@{}}
      \toprule
      $N$ & $P_{VA}$ & $P_{EXPR}$ & $P_{MTL}$ \\
      \midrule
       1 & 0.170 & 0.199 & 0.740 \\
       8 & 0.233 & 0.217 & 0.826 \\
      16 & 0.238 & 0.212 & 0.826 \\
      32 & 0.242 & 0.213 & \textbf{0.829} \\
      \bottomrule
    \end{tabular}
  \end{minipage}%
  \hfill
  \begin{minipage}[t]{0.47\linewidth}
    \centering
    \begin{tabular}{@{}lrrr@{}}
      \toprule
      $T$ & $P_{VA}$ & $P_{EXPR}$ & $P_{MTL}$ \\
      \midrule
      10 & 0.239 & 0.210 & 0.824 \\
      20 & 0.239 & 0.214 & 0.829 \\
      30 & 0.238 & 0.212 & 0.826 \\
      50 & 0.237 & 0.219 & \textbf{0.831} \\
      \bottomrule
    \end{tabular}
  \end{minipage}
\end{table}

Sampling saturates at $N=8$ and all step counts are within 0.007 $P_{MTL}$ of each other.
$N=1$ already outperforms the official baseline ($P_{MTL}=0.45$), consistent with the straight-line trajectory property of rectified flow; we use $N=16$, $T=30$ as the default.

\subsection{PSP: Patch Soft-Pool Feature Aggregation}
\label{sec:app_psp}

Rather than using only the DINOv3 \texttt{[CLS]} token, patch soft-pool (PSP) learns per-channel attention weights over the 196 spatial patch tokens and concatenates the result with the CLS token before projection:
\begin{equation}
  h_i^{\text{PSP}} = \bigl[h_i^{\text{CLS}},\; \textstyle\sum_{p=1}^{196} \sigma_p^{(j)} \cdot h_i^{(p,j)}\bigr]_{j=1}^{d},
  \label{eq:psp}
\end{equation}
where $\sigma^{(j)} = \mathrm{softmax}(w^{(j)})$, weights initialised to uniform.
This doubles the projection input to $2d=768$; only the pooling weights are trained (frozen backbone).

\begin{table}[t]
  \caption{PSP vs.\ AffectFlow baseline. Frozen DINOv3 ViT-S/16, 20 epochs.}
  \label{tab:psp}
  \centering\small
  \setlength{\tabcolsep}{3pt}
  \begin{tabular}{@{}llrrrrrr@{}}
    \toprule
    Config & Decode & CCC-V & CCC-A & $P_{VA}$ & $P_{EXPR}$ & $P_{AU}$ & $P_{MTL}$ \\
    \midrule
    AffectFlow & Flow ($N$=16, $T$=30) & 0.290 & 0.186 & 0.238 & 0.212 & 0.376 & 0.826 \\
    AffectFlow & Det                   & 0.232 & 0.168 & 0.200 & 0.218 & 0.384 & 0.802 \\
    PSP & Flow                 & 0.282 & 0.139 & 0.210 & 0.210 & 0.381 & 0.801 \\
    PSP & Det                  & 0.271 & 0.198 & 0.235 & 0.236 & 0.389 & \textbf{0.859} \\
    \bottomrule
  \end{tabular}
\end{table}

PSP Det reaches $P_{MTL}=0.859$, the best frozen-backbone result without calibration ($+0.033$ over AffectFlow flow), with consistent gains in all three tasks.
PSP Flow underperforms AffectFlow Flow ($0.801$ vs.\ $0.826$), likely because the doubled input dimension requires longer retuning of the flow head; combining PSP with backbone fine-tuning is left as future work.

\section{ViT-B Backbone and Test-Time Augmentation}
\label{sec:app_e24_e26}

Table~\ref{tab:e24_e26} compares the ViT-B backbone and horizontal-flip test-time augmentation (TTA) against the flow-retuned ($\beta{=}1.0$) best checkpoint.

\begin{table}[t]
  \caption{ViT-B backbone and horizontal-flip TTA vs.\ the flow-retuned ViT-S checkpoint. $\dagger$~per-AU calibration applied.}
  \label{tab:e24_e26}
  \centering\small
  \setlength{\tabcolsep}{3pt}
  \begin{tabular}{@{}llrrrrrr@{}}
    \toprule
    Config & Decode & CCC-V & CCC-A & $P_{VA}$ & $P_{EXPR}$ & $P_{AU}$ & $P_{MTL}$ \\
    \midrule
    Flow-retuned ViT-S                  & Det       & 0.387 & 0.263 & 0.325 & 0.296 & 0.453 & 1.073 \\
    Flow-retuned ViT-S$^\dagger$        & Det       & 0.387 & 0.263 & 0.325 & 0.296 & 0.502 & \textbf{1.123} \\
    \midrule
    ViT-B                               & Det       & 0.416 & 0.293 & 0.354 & 0.255 & 0.452 & 1.061 \\
    ViT-B$^\dagger$                     & Det       & 0.416 & 0.293 & 0.354 & 0.255 & 0.507 & 1.116 \\
    Flow-retuned ViT-S+TTA              & Det (avg) & 0.366 & 0.261 & 0.313 & 0.293 & 0.453 & 1.059 \\
    Flow-retuned ViT-S+TTA$^\dagger$    & Det (avg) & 0.366 & 0.261 & 0.313 & 0.293 & 0.504 & 1.111 \\
    \bottomrule
  \end{tabular}
\end{table}

ViT-B (ViT-B/16, 86M parameters) improves valence (CCC-V $0.416$ vs.\ $0.387$) and per-AU calibrated score ($0.507$ vs.\ $0.502$), but degrades $P_{EXPR}$ by $0.041$, yielding $P_{MTL}=1.116$ calibrated, below the ViT-S best ($1.123$).
The VA--EXPR trade-off suggests ViT-B captures finer spatial VA signal but is harder to fine-tune for balanced expression prediction in 20 epochs.
Horizontal-flip TTA hurts all three sub-scores; flip changes asymmetric AU patterns (e.g.\ AU12 corner-of-mouth), explaining the regression.

\section{Additional Ablation Studies}
\label{app:ablations}

\begin{figure}[t]
  \centering
  \includegraphics[width=\linewidth]{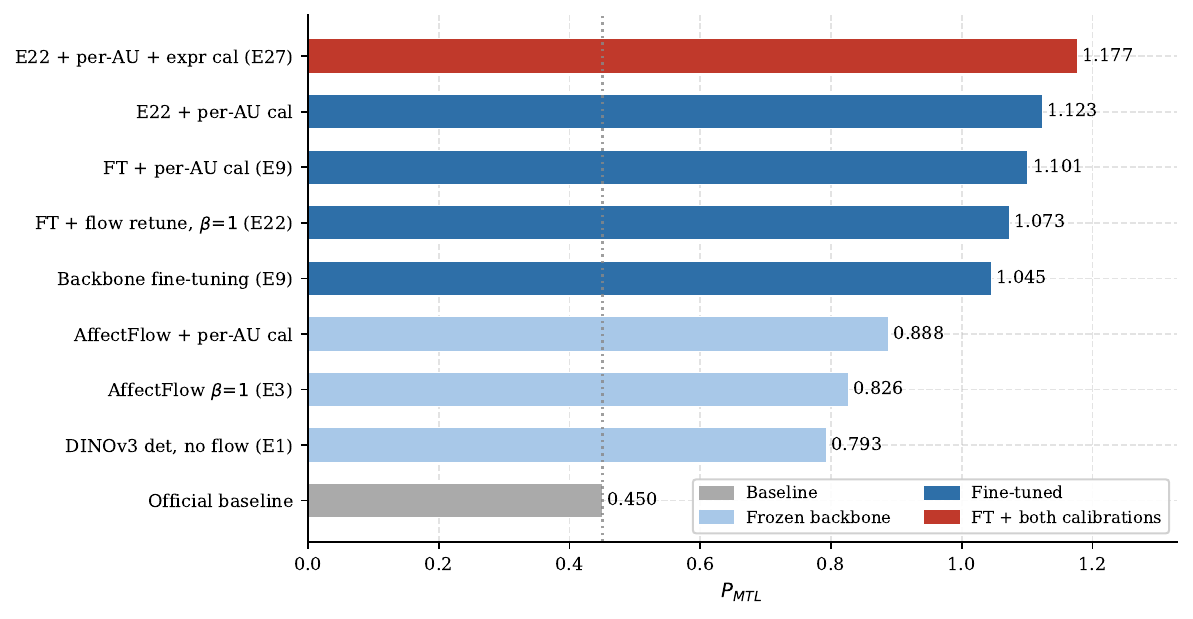}
  \caption{$P_{MTL}$ progression across key configurations.
    Color encodes training stage: frozen backbone (light blue), fine-tuned (dark blue), fine-tuned with both calibrations (red).
    The dotted line marks the official baseline ($P_{MTL}=0.450$).
    Backbone fine-tuning is the largest single jump; post-hoc per-AU and per-class calibration provide a further $+0.132$ at zero retraining cost.}
  \label{fig:ablation_progression}
\end{figure}

This appendix collects ablations that did not improve over the AffectFlow baseline: VA loss weight, global AU threshold sweep, expression-enhancement strategies, label smoothing, and global AU loss weight.

\subsection{VA Loss Weighting}
\label{sec:e7}

Arousal estimation is the weakest component across all runs.
Table~\ref{tab:e7_vaweight} tests whether increasing the VA loss weight $\lambda_{VA} \in \{1, 2, 4\}$ can improve the arousal CCC without sacrificing other tasks.

\begin{table}[t]
  \caption{Effect of VA loss weight $\lambda_{VA}$. Flow decoding with $N=16$ samples and $T=30$ steps.}
  \label{tab:e7_vaweight}
  \centering
  \begin{tabular}{@{}lrrrrrr@{}}
    \toprule
    $\lambda_{VA}$ & CCC-V & CCC-A & $P_{VA}$ & $P_{EXPR}$ & $P_{AU}$ & $P_{MTL}$ \\
    \midrule
    1 & 0.290 & 0.186 & 0.238 & 0.212 & 0.376 & 0.826 \\
    2 & 0.289 & 0.184 & 0.237 & 0.219 & 0.375 & \textbf{0.830} \\
    4 & 0.259 & 0.194 & 0.226 & 0.196 & 0.378 & 0.800 \\
    \bottomrule
  \end{tabular}
\end{table}

Doubling the VA weight ($\lambda_{VA}=2$) yields the best $P_{MTL}$ by improving $P_{EXPR}$ without harming $P_{VA}$, suggesting minor positive cross-task regularization.
At $\lambda_{VA}=4$ the stronger VA supervision improves arousal but reduces expression performance and the overall score, consistent with negative transfer at extreme task weight imbalances~\cite{liu2019mtl}.
The gain at $\lambda_{VA}=2$ is marginal ($+0.004$ $P_{MTL}$) and not consistent across ablations; we retain the default $\lambda_{VA}=1$.

\subsection{Global AU Threshold Sweep}
\label{sec:e8}

The default AU decision threshold is 0.5.
Because AU labels are multi-label and class-imbalanced, we additionally evaluate a simple global threshold sweep on the AffectFlow checkpoint.
Table~\ref{tab:e8_authreshold} shows that a single global threshold does not improve over the default value; lower thresholds increase false positives, while a high threshold of 0.7 sharply reduces AU F1.

\begin{table}[t]
  \caption{Global AU threshold sweep on the AffectFlow checkpoint. Flow decoding uses $N=16$ samples and $T=30$ steps.}
  \label{tab:e8_authreshold}
  \centering
  \begin{tabular}{@{}lrrrrrr@{}}
    \toprule
    AU threshold & CCC-V & CCC-A & $P_{VA}$ & $P_{EXPR}$ & $P_{AU}$ & $P_{MTL}$ \\
    \midrule
    0.3 & 0.289 & 0.192 & 0.240 & 0.212 & 0.359 & 0.811 \\
    0.4 & 0.287 & 0.187 & 0.237 & 0.214 & 0.359 & 0.809 \\
    0.5 & 0.290 & 0.190 & 0.240 & 0.215 & 0.375 & \textbf{0.830} \\
    0.6 & 0.287 & 0.190 & 0.238 & 0.210 & 0.370 & 0.818 \\
    0.7 & 0.290 & 0.188 & 0.239 & 0.213 & 0.174 & 0.627 \\
    \bottomrule
  \end{tabular}
\end{table}

This result confirms that a single global threshold is too coarse for the heterogeneous AU imbalance structure; per-AU calibration (Section~\ref{sec:e14} of the main paper) is necessary.

\subsection{Expression Enhancement Failures}
\label{sec:e10}

Four expression-specific modifications were explored in sequence.
All fail to improve $P_{EXPR}$ over the AffectFlow baseline and consistently degrade $P_{VA}$.

\paragraph{Expression Loss Weight Boost.}
A lightweight proxy for class-balanced training is to increase $\lambda_{EXPR}$, the scalar weight on the expression cross-entropy term, without changing the sampler or loss form.
Table~\ref{tab:e10_exprweight} sweeps $\lambda_{EXPR} \in \{1, 2, 4\}$ on the main AffectFlow checkpoint with flow decoding.

\begin{table}[t]
  \caption{Effect of expression loss weight $\lambda_{EXPR}$. Flow decoding with $N=16$ samples and $T=30$ steps.}
  \label{tab:e10_exprweight}
  \centering
  \begin{tabular}{@{}lrrrrrr@{}}
    \toprule
    $\lambda_{EXPR}$ & CCC-V & CCC-A & $P_{VA}$ & $P_{EXPR}$ & $P_{AU}$ & $P_{MTL}$ \\
    \midrule
    1 & 0.290 & 0.186 & 0.238 & 0.212 & 0.376 & \textbf{0.826} \\
    2 & 0.230 & 0.201 & 0.216 & 0.196 & 0.378 & 0.790 \\
    4 & 0.230 & 0.207 & 0.218 & 0.204 & 0.377 & 0.800 \\
    \bottomrule
  \end{tabular}
\end{table}

Increasing $\lambda_{EXPR}$ does not improve expression macro-F1 and consistently degrades $P_{VA}$.
Because the cross-entropy loss is dominated by majority classes (Neutral, Happiness) regardless of the weight, scaling the loss up merely competes for capacity with the VA task without directing gradient towards minority classes.

\paragraph{Focal Loss for Expression.}
We replace the masked cross-entropy for expression with a focal loss~\cite{lin2017focal} that down-weights easy majority-class samples via the modulating factor $(1-p_t)^\gamma$.
Table~\ref{tab:e11_focal} reports flow-decoding results for $\gamma \in \{1, 2\}$ with all other settings identical to AffectFlow.

\begin{table}[t]
  \caption{Effect of focal loss on the expression head ($\gamma$). Flow decoding with $N=16$ samples and $T=30$ steps.}
  \label{tab:e11_focal}
  \centering
  \begin{tabular}{@{}lrrrrrr@{}}
    \toprule
    $\gamma$ & CCC-V & CCC-A & $P_{VA}$ & $P_{EXPR}$ & $P_{AU}$ & $P_{MTL}$ \\
    \midrule
    0 (AffectFlow baseline) & 0.290 & 0.186 & 0.238 & 0.212 & 0.376 & \textbf{0.826} \\
    1 & 0.252 & 0.192 & 0.222 & 0.193 & 0.379 & 0.794 \\
    2 & 0.256 & 0.190 & 0.223 & 0.192 & 0.379 & 0.794 \\
    \bottomrule
  \end{tabular}
\end{table}

Focal loss does not improve $P_{EXPR}$ and reduces $P_{VA}$ by approximately 0.016.
We attribute this to the relatively small number of valid expression labels (90K training frames) combined with heavy class skew: the focal modulator suppresses gradient from majority classes without providing enough additional signal on rare classes to compensate for the lost VA supervision.

\paragraph{Per-Task Separate Projection Heads.}
To reduce negative transfer through the shared $768\to768$ projection, we give each task head its own projection MLP while keeping a shared projection for the flow network.
Table~\ref{tab:e12_sepheads} compares this design against the shared-projection AffectFlow model.

\begin{table}[t]
  \caption{Effect of per-task separate projection heads. Flow decoding with $N=16$ samples and $T=30$ steps.}
  \label{tab:e12_sepheads}
  \centering
  \begin{tabular}{@{}lrrrrrr@{}}
    \toprule
    Projection & CCC-V & CCC-A & $P_{VA}$ & $P_{EXPR}$ & $P_{AU}$ & $P_{MTL}$ \\
    \midrule
    Shared (AffectFlow)    & 0.290 & 0.186 & 0.238 & 0.212 & 0.376 & \textbf{0.826} \\
    Separate & 0.278 & 0.183 & 0.230 & 0.196 & 0.371 & 0.797 \\
    \bottomrule
  \end{tabular}
\end{table}

Separate projections slightly reduce all three sub-scores, suggesting that the shared representation acts as a useful regularizer rather than a harmful bottleneck in the frozen-backbone regime.

\paragraph{Focal Loss with Separate Projection Heads.}
Combining focal loss with separate projection heads yields the configuration in Table~\ref{tab:e13_combined}: focal loss ($\gamma=2$) plus per-task projections.

\begin{table}[t]
  \caption{Combined focal loss ($\gamma=2$) and separate projection heads. Flow decoding with $N=16$ samples and $T=30$ steps.}
  \label{tab:e13_combined}
  \centering
  \begin{tabular}{@{}lrrrrrr@{}}
    \toprule
    Config & CCC-V & CCC-A & $P_{VA}$ & $P_{EXPR}$ & $P_{AU}$ & $P_{MTL}$ \\
    \midrule
    AffectFlow baseline  & 0.290 & 0.186 & 0.238 & 0.212 & 0.376 & \textbf{0.826} \\
    Combined & 0.277 & 0.182 & 0.230 & 0.199 & 0.371 & 0.800 \\
    \bottomrule
  \end{tabular}
\end{table}

The combined variant improves $P_{EXPR}$ marginally over separate projection heads alone ($+0.003$) but remains well below the AffectFlow baseline.
The consistent failure of these expression-enhancement strategies establishes that expression imbalance requires structural remedies (class-balanced sampling, positive-weighted losses) rather than loss reweighting or architectural changes alone.

\subsection{Label Smoothing}
\label{sec:e18}

Label smoothing distributes a fraction $\epsilon$ of the one-hot probability mass uniformly across expression classes, providing mild regularization against over-confident predictions.
Table~\ref{tab:e18_labelsmooth} sweeps $\epsilon \in \{0.05, 0.10, 0.20\}$.

\begin{table}[t]
  \caption{Label smoothing sweep for the expression head (frozen backbone, deterministic decode).}
  \label{tab:e18_labelsmooth}
  \centering\small
  \setlength{\tabcolsep}{4pt}
  \begin{tabular}{@{}lrrrrrr@{}}
    \toprule
    $\epsilon$ & CCC-V & CCC-A & $P_{VA}$ & $P_{EXPR}$ & $P_{AU}$ & $P_{MTL}$ \\
    \midrule
    0.00 (AffectFlow Det) & 0.232 & 0.168 & 0.200 & 0.218 & 0.384 & 0.802 \\
    0.05          & 0.276 & 0.160 & 0.218 & 0.230 & 0.368 & 0.816 \\
    0.10          & 0.273 & 0.159 & 0.216 & 0.223 & 0.368 & 0.807 \\
    0.20          & 0.265 & 0.172 & 0.218 & 0.193 & 0.391 & 0.802 \\
    \bottomrule
  \end{tabular}
\end{table}

Mild smoothing ($\epsilon=0.05$) raises $P_{EXPR}$ to $0.230$ and CCC-V to $0.276$, but the gain is small and diminishes rapidly: at $\epsilon=0.20$, $P_{EXPR}$ falls below the unsmoothed baseline.
Label smoothing alone is insufficient to close the expression imbalance gap.

\subsection{Global AU Loss Weight}
\label{sec:e21}

As an alternative to per-AU positive weighting, Table~\ref{tab:e21_aulossweight} sweeps the global AU loss weight $\lambda_{AU} \in \{0.5, 1.0, 2.0\}$.

\begin{table}[t]
  \caption{Global AU loss weight sweep (frozen backbone, deterministic decode).}
  \label{tab:e21_aulossweight}
  \centering\small
  \setlength{\tabcolsep}{4pt}
  \begin{tabular}{@{}lrrrrrr@{}}
    \toprule
    $\lambda_{AU}$ & CCC-V & CCC-A & $P_{VA}$ & $P_{EXPR}$ & $P_{AU}$ & $P_{MTL}$ \\
    \midrule
    1.0 (AffectFlow baseline) & 0.232 & 0.168 & 0.200 & 0.218 & 0.384 & 0.802 \\
    0.5               & 0.277 & 0.158 & 0.218 & 0.237 & 0.365 & 0.819 \\
    2.0               & 0.275 & 0.160 & 0.217 & 0.236 & 0.366 & 0.819 \\
    \bottomrule
  \end{tabular}
\end{table}

Neither reducing nor increasing $\lambda_{AU}$ changes $P_{MTL}$ appreciably; both variants plateau at $0.819$.
Per-AU positive weighting (Section~\ref{sec:e19} below) achieves a larger $P_{AU}$ gain ($+0.043$) than any global weight setting, confirming that per-class reweighting is more effective than global loss scaling for AU detection.

\section{Class Distribution Statistics}
\label{app:datastats}

Table~\ref{tab:dataset_stats} provides the full per-class expression counts and AU positive rates referenced throughout the paper.

\begin{table}[t]
  \caption{%
    Class distribution in the s-Aff-Wild2 training and validation splits.
    \emph{Left}: expression counts, frequency, and inverse-frequency weight $w_c = N_\text{valid}/(8 n_c)$ (used in the class-weighted cross-entropy experiment, Section~\ref{sec:e15} of the main paper).
    \emph{Right}: AU positive rate (\% of valid rows with AU=1) and BCE \texttt{pos\_weight} $= n_\text{neg}/n_\text{pos}$ (used in the AU positive-weighting experiment, Section~\ref{sec:e19} below).
  }
  \label{tab:dataset_stats}
  \centering\scriptsize
  \setlength{\tabcolsep}{3pt}
  \begin{minipage}[t]{0.50\linewidth}
    \centering
    \begin{tabular}{@{}lrrrr@{}}
      \toprule
      Class     & Tr count  & Tr\%  & Val\% & $w_c$ \\
      \midrule
      Neutral   & 23{,}976  & 26.5  & 12.2  & 0.47 \\
      Anger     &  4{,}555  &  5.0  &  3.2  & 2.49 \\
      Disgust   &  3{,}168  &  3.5  &  3.7  & 3.58 \\
      Fear      &  3{,}122  &  3.4  &  8.1  & 3.63 \\
      Happiness & 18{,}135  & 20.0  & 24.3  & 0.63 \\
      Sadness   &  7{,}609  &  8.4  & 12.3  & 1.49 \\
      Surprise  &  5{,}228  &  5.8  &  6.5  & 2.17 \\
      Other     & 24{,}852  & 27.4  & 29.8  & 0.46 \\
      \midrule
      Total     & 90{,}645  & \multicolumn{3}{c}{8$\times$ imbalance ratio} \\
      \bottomrule
    \end{tabular}
  \end{minipage}%
  \hfill
  \begin{minipage}[t]{0.47\linewidth}
    \centering
    \begin{tabular}{@{}lrrr@{}}
      \toprule
      AU   & Tr\%$^+$ & Val\%$^+$ & pos\_w \\
      \midrule
      AU1  & 18.1 & 20.6 &  4.5 \\
      AU2  &  8.5 &  9.5 & 10.8 \\
      AU4  & 19.7 & 20.2 &  4.1 \\
      AU6  & 29.6 & 32.8 &  2.4 \\
      AU7  & 40.0 & 49.2 &  1.5 \\
      AU10 & 37.1 & 42.5 &  1.7 \\
      AU12 & 25.9 & 31.0 &  2.9 \\
      AU15 &  2.4 &  2.9 & 41.0 \\
      AU23 &  2.9 &  2.8 & 33.7 \\
      AU24 &  4.8 &  3.0 & 19.7 \\
      AU25 & 68.3 & 76.0 &  0.5 \\
      AU26 & 10.4 & 10.9 &  8.6 \\
      \midrule
      \multicolumn{4}{c}{29$\times$ imbalance (AU25/AU15)} \\
      \bottomrule
    \end{tabular}
  \end{minipage}
\end{table}

\section{Imbalance Remedies: Balanced Sampling and AU Positive Weighting}
\label{app:imbalance}

\subsection{Balanced Expression Sampling}
\label{sec:e16}

\texttt{WeightedRandomSampler} assigns each training frame a sampling probability inversely proportional to its expression class count, rebalancing the effective class distribution without modifying the loss function.
Table~\ref{tab:e16_e17_balance} compares the balanced sampler alone and the balanced sampler combined with class-weighted cross-entropy against the AffectFlow flow baseline.

\begin{table}[t]
  \caption{Expression class-balance strategies. All variants use a frozen backbone (20 epochs); decode mode matches best-checkpoint evaluation.}
  \label{tab:e16_e17_balance}
  \centering\small
  \setlength{\tabcolsep}{4pt}
  \begin{tabular}{@{}llrrrrrr@{}}
    \toprule
    Strategy & Decode & CCC-V & CCC-A & $P_{VA}$ & $P_{EXPR}$ & $P_{AU}$ & $P_{MTL}$ \\
    \midrule
    Baseline  & Flow   & 0.290 & 0.186 & 0.238 & 0.212 & 0.376 & 0.826 \\
    CW-CE       & Det       & --    & --    & 0.210 & 0.239 & 0.385 & 0.834 \\
    Balanced   & Flow   & 0.264 & 0.213 & 0.238 & 0.251 & 0.375 & \textbf{0.864} \\
    CW+Balanced & Det & 0.225 & 0.162 & 0.193 & 0.246 & 0.394 & 0.833 \\
    \bottomrule
  \end{tabular}
\end{table}

The balanced sampler raises $P_{EXPR}$ to $0.251$ ($+0.039$ over AffectFlow flow) and lifts $P_{MTL}$ to $0.864$, the best frozen-backbone result for expression.
Combining balanced sampling with class-weighted cross-entropy gives $P_{EXPR}=0.246$ under deterministic decode but $P_{VA}$ drops substantially ($0.193$ vs.\ $0.238$), indicating that the combined strategy over-suppresses gradient on majority classes at the cost of valence-arousal prediction.

\subsection{BCE Positive Weighting for AUs}
\label{sec:e19}

We set each AU's BCE positive weight to $n_\text{neg}/n_\text{pos}$ computed from the training set (Table~\ref{tab:dataset_stats}).
One variant applies positive weighting to the AU head alone; another further adds class-weighted expression cross-entropy.
Table~\ref{tab:e19_e20_posweight} reports deterministic predictions throughout: flow decoding degrades both runs ($P_{MTL}=0.777$ for AU positive weighting alone and $0.745$ for the combined variant) because the reweighted loss shifts predicted probability distributions incompatibly with the rectified-flow sampler.

\begin{table}[t]
  \caption{AU BCE positive weighting (frozen backbone, deterministic decode). Flow decoding degrades under the reweighted loss.}
  \label{tab:e19_e20_posweight}
  \centering\small
  \setlength{\tabcolsep}{4pt}
  \begin{tabular}{@{}lrrrrrr@{}}
    \toprule
    Config & CCC-V & CCC-A & $P_{VA}$ & $P_{EXPR}$ & $P_{AU}$ & $P_{MTL}$ \\
    \midrule
    AffectFlow baseline, Det      & 0.232 & 0.168 & 0.200 & 0.218 & 0.384 & 0.802 \\
    AU pos-wt        & 0.261 & 0.167 & 0.214 & 0.237 & 0.427 & 0.878 \\
    AU pos-wt + CW-CE   & 0.259 & 0.163 & 0.211 & 0.240 & 0.439 & \textbf{0.890} \\
    \bottomrule
  \end{tabular}
\end{table}

AU positive weighting raises $P_{AU}$ from $0.384$ to $0.427$ ($+0.043$) at training time.
Adding expression class weighting further lifts $P_{AU}$ to $0.439$, yielding $P_{MTL}=0.890$, the best frozen-backbone result without calibration, marginally surpassing AffectFlow after per-AU calibration ($0.888$).

\section{Kitchen-Sink Fine-Tuning}
\label{app:e23}
\label{sec:e23}

We test whether the most effective frozen-backbone interventions, class-weighted cross-entropy and per-AU BCE positive weighting, transfer additively to the fine-tuned setting.
Both are applied jointly during the same low-LR fine-tuning run as the plain fine-tuning experiment (Section~\ref{sec:e9} of the main paper).
Table~\ref{tab:e23_kitchensink} reports deterministic predictions with and without per-AU calibration.

\begin{table}[t]
  \caption{Kitchen-sink fine-tuning: FT + class-weighted CE + AU pos-weight, deterministic decode. $^\dagger$ per-AU calibration applied.}
  \label{tab:e23_kitchensink}
  \centering\small
  \setlength{\tabcolsep}{3pt}
  \begin{tabular}{@{}lrrrrrr@{}}
    \toprule
    Config & CCC-V & CCC-A & $P_{VA}$ & $P_{EXPR}$ & $P_{AU}$ & $P_{MTL}$ \\
    \midrule
    Plain FT, Det                       & 0.352 & 0.285 & 0.318 & 0.285 & 0.441 & 1.045 \\
    Kitchen-sink FT+CW+pos-wt, Det            & 0.319 & 0.223 & 0.271 & 0.288 & 0.483 & 1.041 \\
    Kitchen-sink FT+CW+pos-wt$^\dagger$, Det  & 0.319 & 0.223 & 0.271 & 0.288 & 0.503 & 1.061 \\
    \bottomrule
  \end{tabular}
\end{table}

The kitchen-sink combination does not improve $P_{MTL}$ over plain fine-tuning without calibration ($1.041$ vs.\ $1.045$), despite raising $P_{AU}$ from $0.441$ to $0.483$.
The AU gain is offset by a $P_{VA}$ drop from $0.318$ to $0.271$: class-weighted cross-entropy reduces gradient allocation for the VA task when the backbone is jointly optimized, consistent with the pattern seen when combining balanced sampling with class-weighted CE (Section~\ref{sec:e16}).
Expression macro-F1 ($0.288$) is nearly identical to plain fine-tuning ($0.285$), confirming that AU positive weighting absorbs gradient competition without a meaningful expression benefit.
After per-AU calibration, the kitchen-sink variant reaches $P_{MTL}=1.061$, below the best overall flow-retuned ($\beta{=}1.0$) calibrated result ($1.123$).
These results indicate that once the backbone is fine-tuned, frozen-backbone class-reweighting strategies add little value and can harm $P_{VA}$; flow retuning is a more effective path.

\section{Per-Class and Per-AU F1 Breakdowns}
\label{app:perau}

\begin{figure}[t]
  \centering
  \includegraphics[width=\linewidth]{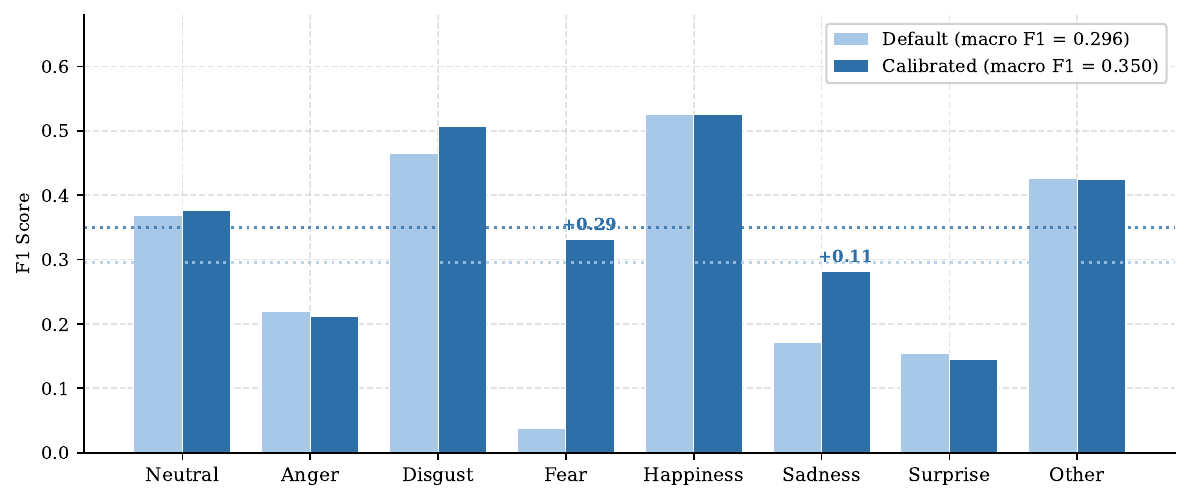}
  \caption{Per-class expression F1 before (light) and after (dark) threshold calibration on the fine-tuned+flow retune ($\beta{=}1.0$) checkpoint.
    Dotted horizontal lines mark the macro F1 at each setting.
    Fear benefits most ($0.038\to0.331$, $+0.293$), followed by Sadness ($0.171\to0.282$, $+0.111$);
    macro F1 rises from $0.296$ to $0.350$.}
  \label{fig:expr_calibration}
\end{figure}

Table~\ref{tab:e15_perclass} reports the full per-class expression F1 for the AffectFlow and fine-tuned models referenced in Section~\ref{sec:e15} of the main paper, and Table~\ref{tab:e15_perau} reports the per-AU F1 scores before and after per-AU threshold calibration for the fine-tuned checkpoint.
Table~\ref{tab:e27_exprcal} and Figure~\ref{fig:expr_calibration} report the per-class expression calibration results from Section~\ref{sec:e27} of the main paper.

\begin{table}[t]
  \caption{Expression F1 (\%) per class for AffectFlow (flow, $N$=16) and the fine-tuned backbone (deterministic).}
  \label{tab:e15_perclass}
  \centering\small
  \setlength{\tabcolsep}{4pt}
  \begin{tabular}{@{}lrr@{}}
    \toprule
    Class     & AffectFlow Flow & Fine-tuned Det \\
    \midrule
    Neutral   & 35.7 & 41.1 \\
    Anger     &  1.1 & 25.3 \\
    Disgust   & 11.8 & 37.0 \\
    Fear      &  6.3 &  6.9 \\
    Happiness & 55.2 & 52.2 \\
    Sadness   &  4.3 &  7.1 \\
    Surprise  &  2.8 & 10.2 \\
    Other     & 52.7 & 48.6 \\
    \midrule
    Macro-F1  & 21.2 & 28.5 \\
    \bottomrule
  \end{tabular}
\end{table}

\begin{table}[t]
  \caption{Per-AU F1 (\%) for the fine-tuned model (deterministic) at threshold 0.5 and after per-AU calibration.}
  \label{tab:e15_perau}
  \centering\small
  \begin{tabular}{@{}lrr@{}}
    \toprule
    AU   & F1@0.5 & F1@cal \\
    \midrule
    AU1   & 51.3 & 55.2 \\
    AU2   & 41.7 & 44.3 \\
    AU4   & 47.9 & 53.4 \\
    AU6   & 50.4 & 55.9 \\
    AU7   & 66.0 & 73.4 \\
    AU10  & 66.7 & 70.7 \\
    AU12  & 64.8 & 65.3 \\
    AU15  &  0.0 & 10.8 \\
    AU23  &  0.5 & 18.9 \\
    AU24  & 22.1 & 23.3 \\
    AU25  & 85.3 & 87.8 \\
    AU26  & 33.0 & 37.5 \\
    \midrule
    Mean  & 44.1 & 49.7 \\
    \bottomrule
  \end{tabular}
\end{table}

\begin{table}[t]
  \caption{Per-class expression F1 before (default $\arg\max$) and after per-class weight calibration on the fine-tuned+flow retune ($\beta{=}1.0$) deterministic predictions. Macro-F1 improves from $0.296$ to $0.350$.}
  \label{tab:e27_exprcal}
  \centering\small
  \setlength{\tabcolsep}{5pt}
  \begin{tabular}{@{}lcc@{}}
    \toprule
    Expression & Default F1 & Calibrated F1 \\
    \midrule
    Neutral   & 0.368 & 0.377 \\
    Anger     & 0.219 & 0.212 \\
    Disgust   & 0.465 & 0.507 \\
    Fear      & 0.038 & 0.331 \\
    Happiness & 0.526 & 0.525 \\
    Sadness   & 0.171 & 0.282 \\
    Surprise  & 0.154 & 0.145 \\
    Other     & 0.426 & 0.425 \\
    \midrule
    Macro-F1 ($P_{EXPR}$) & 0.296 & \textbf{0.350} \\
    \bottomrule
  \end{tabular}
\end{table}

AU15 and AU23, the two rarest AUs (positive rate 2.4\% and 2.9\%), have near-zero F1 at the default threshold but recover to $10.8\%$ and $18.9\%$ after per-AU calibration, confirming that the model has learned discriminative signal that the imbalanced training objective systematically suppresses.

\end{document}